\documentclass{article} 
\usepackage{baichuan}

\usepackage[utf8]{inputenc} 
\usepackage[T1]{fontenc}    
\usepackage{hyperref}       
\usepackage{url}            
\usepackage{booktabs}       
\usepackage{amsfonts}       
\usepackage{nicefrac}       
\usepackage{microtype}      
\usepackage{xcolor}         
\usepackage{multirow}
\usepackage{multicol}
\usepackage{amsmath}
\usepackage{graphicx}
\usepackage{amsmath}

\newtheorem{lemma}{Lemma}

\usepackage{subcaption}
\usepackage{utfsym}
\usepackage{svg}
\usepackage{array}
\usepackage{graphicx}
\usepackage{fancyhdr}
\usepackage{tablefootnote}
\usepackage{footnote}
\usepackage{wrapfig}
\usepackage{threeparttable}

\title{ShortGPT: Layers in Large Language Models are More Redundant Than You Expect }

\author{%
	\quad 	Xin Men${*}$ \\ 
	Baichuan Inc. \\
	\And
	Mingyu Xu${*}$ \\
	Baichuan Inc. \\
	\And
	Qingyu Zhang\thanks{Equal contribution} \\
	\quad\quad ISCAS \\
	\And
	Bingning Wang \thanks{Corresponding author, \texttt{daniel@baichuan-inc.com}} \quad\quad\\
	\quad  Baichuan Inc. \\ 
	\AND
	Hongyu Lin\quad\quad\quad\\
	\quad ISCAS \\
	\And
	Yaojie Lu\quad\quad\quad\\
	\quad ISCAS \\
	\And
	Xianpei Han \quad\quad\quad \\ 
	\quad ISCAS  \\
	\And
	\And
	Weipeng Chen \\
	Baichuan Inc. \\
}

\colmfinalcopy

\begin{document}
	\maketitle
	\begin{abstract} \label{lab:abstract}
		As Large Language Models (LLMs) continue to advance in performance, their size has increased significantly, with current LLMs containing billions or even trillions of parameters.  In this study, we identify notable redundancy across the layers of LLMs, where some layers contribute minimally to overall network functionality. To quantify this, we introduce a metric called Block Influence (BI) which use the similarity between layer's input and output to measure the importance of each layer. Based on the observation of layer redundancy, we propose a straightforward pruning method: layer removal, which eliminates redundant layers based on their BI scores. Our approach, termed ShortGPT, demonstrates superior performance over previous state-of-the-art pruning methods.  Moreover, ShortGPT is orthogonal to quantization-like methods, enabling further reduction in parameters and computation. The ability to achieve better results through simple layer removal, as opposed to more complex pruning techniques, suggests a high degree of redundancy across layers, not only in transformer models but also in non-transformer models. We hope this work will contribute to future research in LLM compression.
	\end{abstract}
	\section{Introduction}\label{lab:intro}
	The field of large language models (LLMs) has witnessed rapid development recently, with LLMs achieving impressive performance across various domains. Guided by the scaling laws identified in prior work \citep{kaplan2020scaling,hoffmann2022training}, current LLM research tend to increase model parameters to boost performance. As a result, modern LLMs, which can comprise billions to trillions of parameters, require significant hardware resources for deployment, creating substantial barriers to their practical use.       
	
	To mitigate the hardware demands of large models, model compression techniques have become a critical area of focus \citep{zhu2023survey}. These techniques are generally divided into quantization \citep{liu2021post,gholami2022survey,dettmers2022llm,dettmers2024qlora} and pruning\citep{lecun1989optimal,han2015learning,frantar2023massive}. Quantization reduces the precision of model parameters, but its effectiveness often requires specific hardware support. In contrast, pruning method removes redundant parameters to decrease the model's size and computation, offering a more flexible and hardware-agnostic approach. Despite its advantages, many existing pruning methods are  complex; for example, some require gradient information \citep{ma2024llm}, which limits their practicality.
	
	
	In this paper, we focus on the issue of layer redundancy in LLMs and propose a novel approach for simplifying these models. We introduce \textbf{Block Influence (BI)}, a metric that quantifies how much the hidden state changes after passing through each layer, providing a more direct measure of a layer's importance. Leveraging this insight, we propose a simple yet effective pruning method \textbf{ShortGPT}, which identifies and removes layers with lower BI scores, significantly reducing model size without sacrificing much performance. 
	
	To evaluate our approach, we conducted evaluation across comprehensive benchmarks.  Our experiments revealed that our method exhibits a smaller performance decrement compared to the previous methods. For instance,  removing 10 layers (25\% of the total 40 layers) from the LLaMA 2-13B model resulted in only a slight drop in performance on the MMLU benchmark \citep{hendrycks2020measuring}, from 55.0 to 52.2. Our findings highlight substantial redundancy in current LLMs and suggest potential avenues for improving the efficiency of model training by reducing inherent redundancy in the future.
	
	The main contributions of our paper are summarized as follows:
	\begin{itemize}
		\item We analyze the redundancy in large language models (LLMs) and find that they exhibit significant redundancy at the layer level. This finding inspire us to prune LLMs by simply removing redundant layers. 
		\item We propose a metric called Block Influence (BI) as an  indicator of layer importance. Based on BI,  our layer removal method maintains approximately 90\% performance while reducing approximately 25\% of  parameters, outperforming previous state-of-the-art methods.
		\item Furthermore, we demonstrate that our layer pruning approach is orthogonal to quantization methods, meaning it can be combined with quantization techniques to further reduce the deployment overhead of LLMs.
	\end{itemize}	
	
	\begin{figure}[t]
		\centering
		\begin{subfigure}[t]{0.45\textwidth}
			\centering  
			\includegraphics[width=\textwidth]{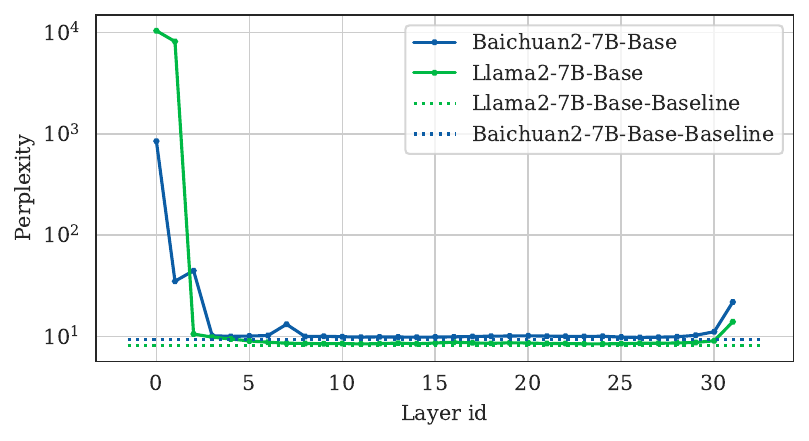}
			\caption{Perplxity}
			\label{fig:overall_illustartion_llama2:1}
		\end{subfigure}
		\hfill
		\begin{subfigure}[t]{0.45\textwidth}
			\centering  
			\includegraphics[width=\textwidth]{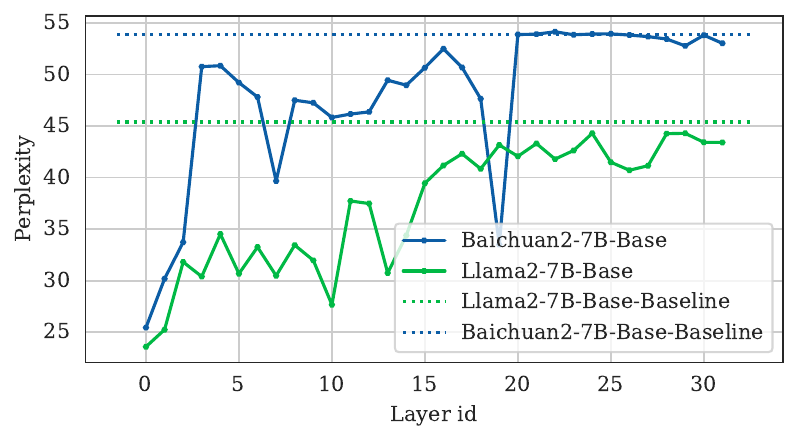}
			\caption{MMLU}
		\end{subfigure}
		\caption{Performance of removing certain layer from LLMs. We can see that certain layers are redundant, and their removal results in minimal performance degradation. }    
		\label{fig:background-redundancy}    
	\end{figure}
	\section{Motivation}
	\subsection{Background}
	
	The predominant LLMs are primarily based on the Transformer architecture \citep{vaswani2017attention}, with the pre-norm configuration being the most commonly adopted, as in models like LLaMA \citep{touvron2023llama}. The pre-norm configuration, where layer normalization is applied before the self-attention and feed-forward layers, offers several advantages such as faster convergence, improved training stability, and better scalability for deeper networks \citep{xiong2020layer, liu2020understanding, wang2024deepnet}. Due to these benefits, the pre-norm approach has been adopted even in non-transformer models, such as  Mamba \citep{gu2023mamba} and RWKV \citep{peng2023rwkv}. For the sake of simplicity in descriptions, our analysis primarily focuses on the Transformer architecture, though we extend our experiments to non-Transformer structures in Section \ref{sec:non-transformer}. 
	
		\begin{figure}[h]
			\centering  
			\includegraphics[width=0.8\textwidth]{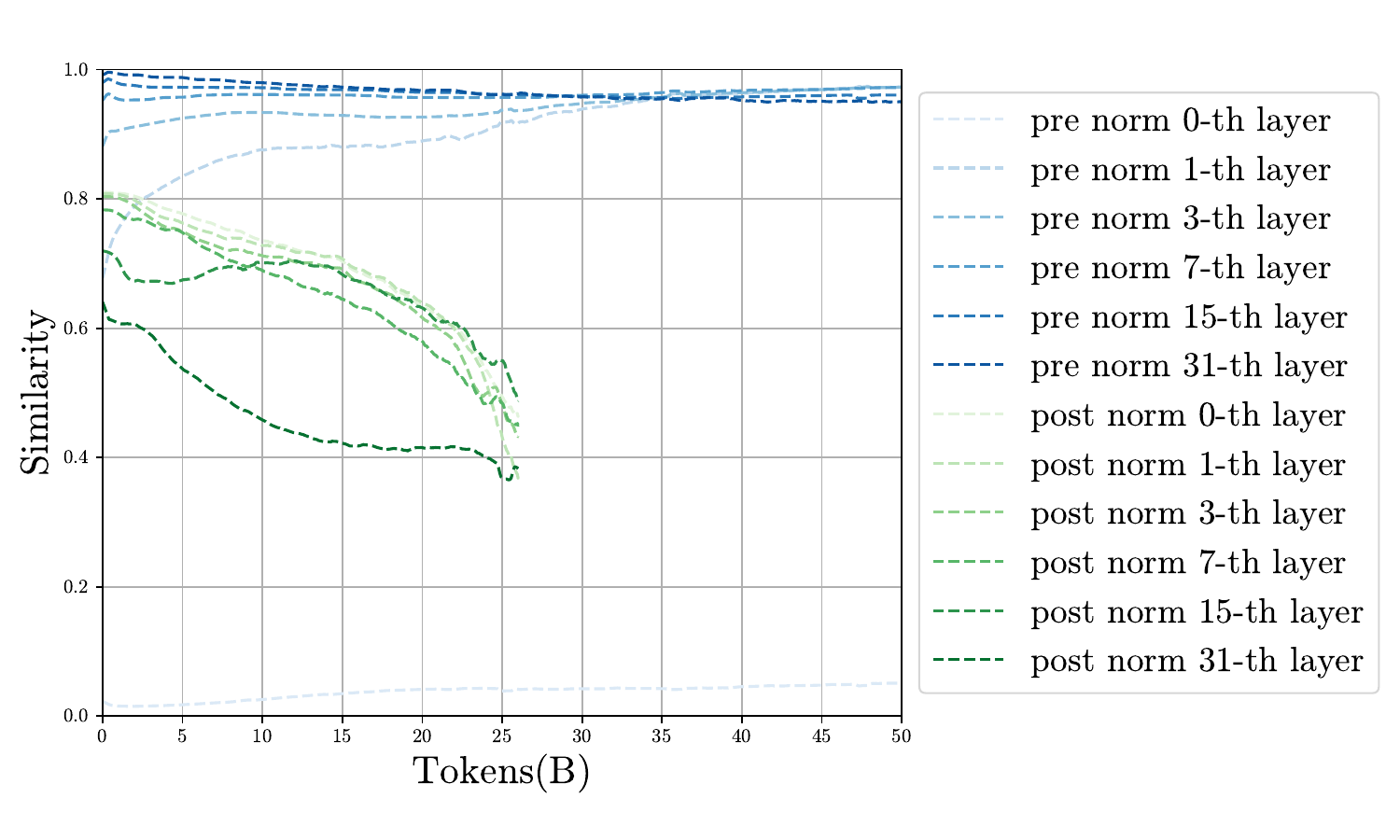}
			\caption{The cosine similarity  between a layer's input and output during the training process. The horizontal axis (X-axis) represents the number of training tokens, while the vertical axis (Y-axis) depicts the degree of similarity. Notably, the model employing post-normalization exhibits divergence after approximately $\sim$26B tokens of training. Training setting is provided in \ref{appendix:post and pre}.}
			\label{fig:background-similarity}    
		\end{figure}
		
		However, we observe that when pre-norm is adopted, the similarity between the input and output of transformer layers tends to be higher, as illustrated in Figure \ref{fig:background-similarity}. This high similarity indicates that certain layers induce minimal changes to the hidden states, suggesting they contribute little to the model’s overall function. A detailed mathematical explanation for this phenomenon is provided in Appendix \ref{appendix:math}. Which suggests that the deep layers of the model with pre-norm might not play a critical role in the overall function, and that \textbf{the layers in large language models could be more redundant than expected}, which motivates the layer-removal based pruning method we explore in the next section.

		\subsection{Layer redundancy}\label{layerredundancy}
		\begin{wraptable}{r}{0.45\textwidth}\label{tab:last layer}
				\centering
				\caption{Ablation of removing FFN and Attention of Llama2-7B-Base. We sample 100 instances from PG19 \citep{rae2019compressive} to calculate PPL.}
				\label{tab:last_layer}
				\begin{tabular}{@{}ll@{}}
					\toprule
					\textbf{Delete} & \textbf{PPL} \\ \midrule
					None & 7.60\\
					The whole last layer & 13.37 \\
					Attention of the last layer & 7.65 \\
					FFN of the last layer & 12.35 \\ \bottomrule
				\end{tabular}
		\end{wraptable}
		As discussed in the previous section, we speculate that the LLMs exhibit layer redundancy. To verify this, we assess the performance degradation caused by removing individual layers of two popular models, Llama2-7B-Base \citep{touvron2023llama}, an English based LLMs, and Baichuan2-7B-Base \citep{yang2023baichuan} which is mainly focused on Chinese. Figure \ref{fig:background-redundancy} confirms our speculation, which reveals that some layers do not play a crucial role in LLMs, causing little degradation when omitting them individually. Moreover, this redundancy is primarily manifested in the middle to later layers of the network, with the initial layers and the last layer often being more critical. Notably, we found the last layer to be particularly important, aligning with findings from LLM Pruner \citep{ma2024llm}. This observation contradicts our mathematical explanation in Appendix \ref{appendix:math} which suggests that deeper layers tend to be more redundant. We posit that this discrepancy arises because the final FFN effectively functions as part of the token classifier and should be considered in conjunction with the language model head.To verify our hypothesis, we conducted further investigation, detailed in Table \ref{tab:last_layer}. The results show that within the last layer, the FFN component is crucial, while the Attention module is less significant. This finding supports our interpretation of the final layer's importance.


		\section{Methodology}
		In this section, we present the methodological framework of our layer removal approach for LLMs, elucidating the underlying principles and techniques employed. We begin by introducing Block Influence (BI), a novel metric designed to assess the hidden states transformation of each layer. Leveraging BI, we then detail our layer removal method.

		\subsection{Layer importance} \label{method:layerimportacne}
\begin{figure}[t]
			\centering
			\begin{subfigure}[t]{0.49\textwidth}
				\centering  
				\includegraphics[width=\textwidth]{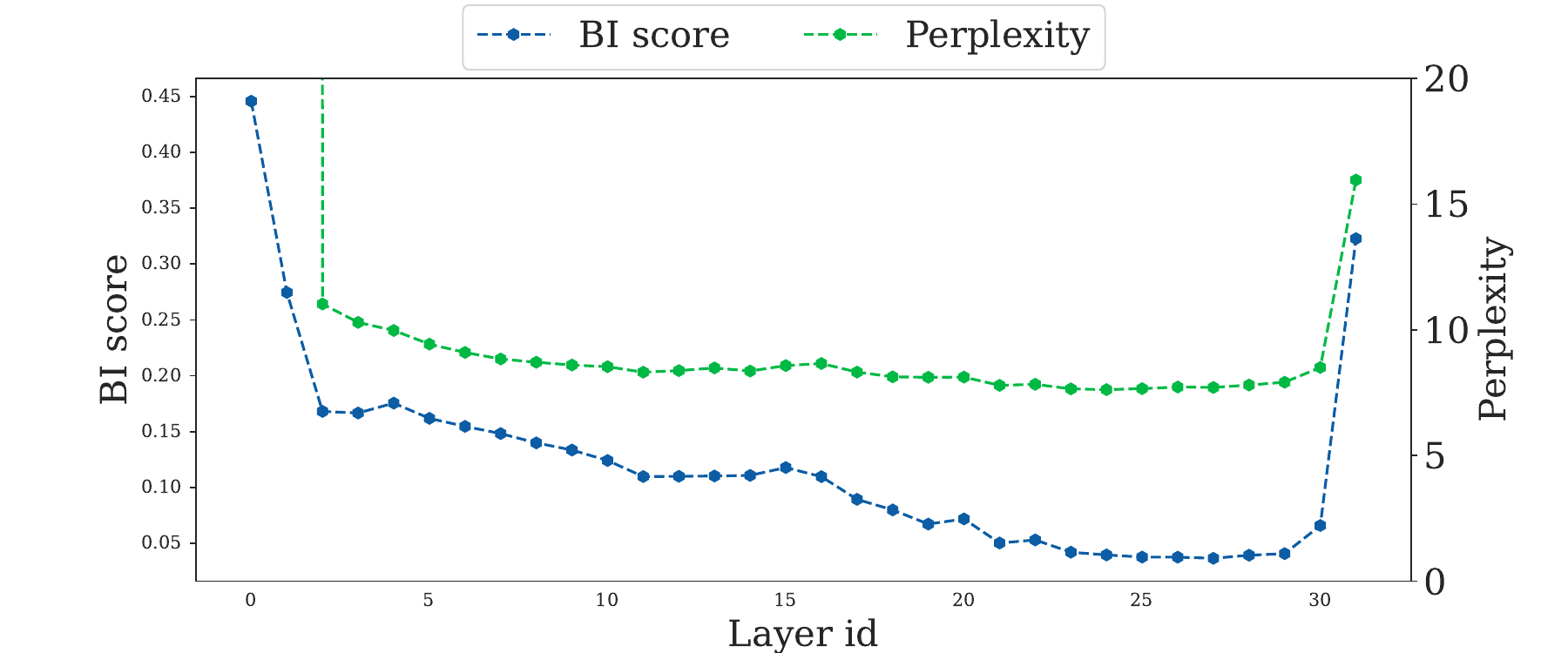}
				\caption{Llama2 7B}
			\end{subfigure}
			\hfill
			\begin{subfigure}[t]{0.49\textwidth}
				\centering  
				\includegraphics[width=\textwidth]{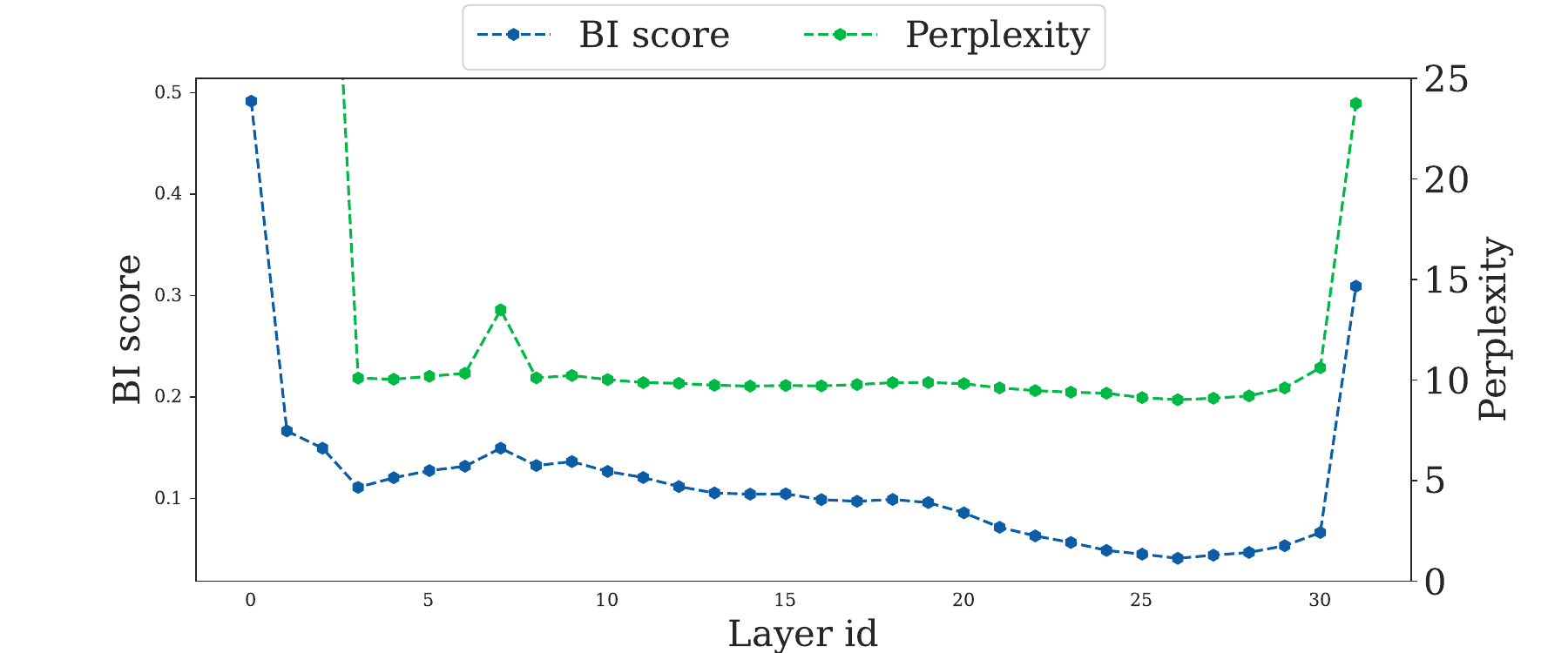}
				\caption{Baichuan2 7B }
			\end{subfigure}
			\caption{The BI score of a layer and the PPL after removing the layer. }    
			\label{fig:bi_ppl}    
		\end{figure}

		As outlined in the preceding section, the layers of LLMs exhibit redundancy, with varying degrees of redundancy across different layers. To capture this, we introduce a new metric, Block Influence (BI), to measure the degree of transformation performed by each layer.   The BI score of $i^{th}$ layer can be calculated as follows: 
		
		
		\begin{align}
			\text{BI}_i = 1 - \mathbb{E}_{X,t} \frac{X_{i,t}^TX_{i+1,t}}{||X_{i,t}||_2||X_{i+1,t}||_2},
		\end{align}
		where $X_{i,t}$ means the $t^{th}$ row of hidden states of $i^{th}$ layer. Lower BI score imply that $X_i$ and $X_{i+1}$ exhibit high cosine similarity, suggesting that the layer makes minimal transformations to the hidden states and is therefore less important. We plot the BI scores of a single layer and the PPL after removing it separately, as shown in the Figure \ref{fig:bi_ppl}. The results demonstrate a positive correlation between the BI score and the importance of a layer.
		
		\subsection{Layer Removal}
		Our goal is to obtain a pruned model that remains as close as possible to the original model. Since an LLM functions as a series of transformations applied to hidden states across its layers and we can determine the importance of each layer, we propose a straightforward pruning method: layer removal, which we refer to as ShortGPT. We delete certain layers in LLMs based on BI score. First of all, we construct a calibration set, which is a set of unlabelled text samples such as PG19 \citep{rae2019compressive}. 
		Then we collect the hidden states of each layer during inference on these samples. Next, we calculate the BI score based on the collected hidden states. Finally, we sort layers in ascending order according to the BI, and delete the layers with the lower BI score. The number of layers to be deleted can vary to trade off the speed and performance. The details of our layer removal setting can be found in Appendix \ref{appendix:remove_strategy}.
		
		\section{Experiments}\label{exp}
		\subsection{Experimental Setup} \label{label:exp_setup}
		\paragraph{Models.}To validate the effectiveness of our method, we conducted experiments on existing popular open-source language models, including Llama2-7B \citep{touvron2023llama}, Llama2-13B, Baichuan2-7B, and Baichuan2-13B. They are all large language models based on the decoder-only Transformer architecture. LLaMA 2 was trained on more than 2 trillion tokens. Baichuan-series was mainly trained in Chinese and its 13-Billion model replaced the RoPE \citep{su2024roformer} positional embedding with ALiBi \citep{press2021train}. 

		\paragraph{Benchmarks.} In order to comprehensively evaluate the changes in the ability of large language models before and after pruning, we conducted comprehensive evaluation from five aspect: \textbf{Reasoning}: CMNLI \citep{li2024cmmlu}, HellaSwag (HeSw) \citep{zellers2019hellaswag}, PIQA \citep{bisk2020piqa}. \textbf{Language}: CHID \citep{zheng2019chid},  WSC (Levesque et al., 2012). \textbf{Knowledge}: CommonSenseQA (CoQA) \citep{reddy2019coqa}, BoolQ \citep{clark2019boolq}. \textbf{Examination}: MMLU \citep{hendrycks2020measuring}, CMMLU \citep{li2024cmmlu}. \textbf{Understanding}: Race-High/Middle (H/M) \citep{lai2017race}, XSum \citep{hasan2021xl}, C3 \citep{sun2020investigating} and PG19 \citep{rae2019compressive}. For more details, please refer to Appendix \ref{appendix:benchmark}
		
		\paragraph{Baselines.} 
		
		To evaluate the effectiveness of our method, we compared several structured pruning methods for large language models, including:
		
		\textbf{1) LLMPru} \citep{ma2024llm}, which  adopts structural pruning that selectively removes non-critical coupled structures based on gradient information, maximally preserving the majority of the LLM’s functionality. LLMPru. applies post training to the pruned model, but for fair comparison, we do not apply post training to it.
		
		\textbf{2) SliceGPT} \citep{ashkboos2024slicegpt}, which is a post-training sparsification scheme that replaces each weight matrix with a smaller matrix, reducing the embedding dimension of the network. Specifically, they applied PCA to the hidden representation from shallow to deep layers, and incorporated the dimension reduction matrix into existing network parameters.
		
		\textbf{3) LaCo} \citep{yang2024laco}, which is a pruning method for large language models based on reducing layers. LaCo gradually merges similar layers from deep to shallow and sets a threshold to avoid continuously merging too many layers.
		
		For our evaluation, we use PG19 for layer importance and perplexity calculation. The models, baselines and evaluate benchmarks is the same as LaCo. 
		
		\renewcommand\arraystretch{1.3} 
		\begin{table}[t]
			\tiny
			\setlength{\tabcolsep}{1.6pt}
			\caption{Comparison of pruning methods on multiple natural language benchmarks. The results of LLMPrun., SliceGPT and LaCo are reported from LaCo. The last column reports the relative performance retention.}
			\label{tab:llm_comparison_all}
			\centering
			\begin{tabular}{c|c|c|ccccccccccccc|cc}
				\hline
				\hline
				\multirow{2}{*}{LLM} & \multirow{2}{*}{Method}  & \multirow{2}{*}{Ratio}& \multicolumn{13}{c|}{Benchmarks}& \multirow{2}{*}{Ave.} & \multirow{2}{*}{Per.} \\  
				& & & CMNLI & HeSw&PIQA&CHID&WSC&CoQA&BoolQ&Race-H&Race-M&XSum&C3 &MMLU & CMMLU& &\\
				\hline
				\multirow{5}{*}{Llama2-7B} & Dense  & 0.00\% &32.99 &71.26 &77.91  &41.66 &50.00  &64.62 &71.62 &35.71  &34.19 &19.40 &43.56&45.39& 32.92 &47.78 &100.00
				\\
				& LLMPrun. & 27.0\% & \textbf{34.33} &\textbf{56.46}& \textbf{71.22} &25.25 &36.54  &42.51 &55.20 &22.56 &22.35 &11.51 &25.64& 23.33  &  25.25 &34.78&72.79\\
				& SliceGPT  & 26.4\%& 31.70 &50.27 &66.21 &20.79 &36.54 & 41.36& 38.32&  21.07 &21.66& 4.89& \textbf{39.78}&28.92 & 25.37 &32.84&68.73   \\
				& LaCo  & 27.1\%& 34.43 &55.69& 69.80 &\textbf{36.14} &40.38  &45.70 &64.07 &22.61& 23.61 &\textbf{15.64} &39.67&  26.45& 25.24&38.41& 80.39\\
				& ShortGPT  & 27.1\%& 32.95	&53.02	&66.43	&24.68	&\textbf{52.46}	&\textbf{47.99}		&\textbf{74.71}&	\textbf{32.25}&	\textbf{35.17}&	0.67	&39.62&\textbf{43.96} &	\textbf{32.25}&\textbf{41.24}&\textbf{86.31} \\
				\hline
				\multirow{5}{*}{Llama2-13B} & Dense &0.00\%&32.99 	&74.78	&79.71	&47.35	&50.00	&66.91 &82.39 &57.95	&60.38	&23.45	&47.51&55.00 &38.40&55.14 &100.00  \\
				& LLMPrun. &24.4\%&\textbf{33.03} &\textbf{67.76} &\textbf{76.66} &35.64 &40.38  &50.86 &56.42  &22.47 &22.08 &\textbf{19.17} &32.33 &25.21& 24.71&38.97&70.67\\
				& SliceGPT &23.6\%&29.82 &55.71 &69.04 &19.31 &36.54  &47.26 &37.86 &23.41 &24.03 &5.27 &41.92 &  37.14& 25.79&34.85&63.20\\
				& LaCo &24.6\%&32.86 &64.39 &63.20 &\textbf{40.10} &\textbf{52.88}  &52.66&\textbf{63.98} &54.49 &56.55 &14.45 &44.93&  45.93& 32.62 &47.62 &86.36\\
				& ShortGPT  &24.6\%&33.00	&66.64&	73.45&	36.61	&50.00		&\textbf{58.64}&	62.48	&\textbf{58.35}	&\textbf{60.17}&	17.59	&\textbf{46.90}&\textbf{54.69}	&\textbf{38.38}&\textbf{50.53}& \textbf{91.64} \\
				\hline
				\multirow{5}{*}{Baichuan2-7B} & Dense &0.00\%&33.37 &67.56	&76.17	&85.56	&50.00	 &63.14 &74.10 &52.63	&51.04	 &20.82	&64.55& 53.87	& 56.95 &57.67&100.00 \\
				& LLMPrun. &24.2\%&32.28 &53.66 &\textbf{71.82} &69.80 &\textbf{53.85} &\textbf{47.83} &61.19 &21.96 &22.28 &\textbf{15.98} &41.64 &  24.93 & 25.69 &41.76& 72.41 \\
				& SliceGPT &22.2\%&32.07 &25.29 &50.33 &14.85 &36.54  &19.57 &39.30 &23.53 &22.49 &0.00 &26.58&  25.18 &25.25&26.23&45.48 \\
				& LaCo &24.2\%&33.00 &52.28 &68.50 &\textbf{76.24} &42.31 &47.26 &56.15 &28.99 &27.72 &12.03 &50.85& 31.53 &31.24&42.93 & 74.44 \\
				& ShortGPT &24.2\%&\textbf{33.30} 	&\textbf{56.96}	&67.68	&65.63	&50.00	 &46.70 &\textbf{67.83} &\textbf{53.26} &\textbf{46.76}&0.04 &\textbf{56.33}& \textbf{45.77} &	\textbf{47.87} &\textbf{49.08}&\textbf{85.10} \\
				\hline
				\multirow{5}{*}{Baichuan2-13B} & Dense&0.00\%& 33.21 	&71.10	&78.07	&86.51	&50.00 &65.6 &77.89 &67.27	&68.94	 &25.02	&65.64 &  59.50 &61.30&62.31&100.00   \\
				& LLMPrun. &24.3\%&\textbf{33.80} &53.57 &\textbf{71.82} &72.77 &37.50  &38.82 &56.54 &21.17 &21.61 &13.67 &39.89&  23.19 & 25.18&39.20&62.91 \\
				& SliceGPT &22.8\%&32.07 &25.85 &51.03 &10.40 &36.54  &18.02 &37.83 &21.56 &21.52 &0.00 &24.99&  22.95 & 25.26&25.23& 40.49\\
				& LaCo &24.7\%&33.03 &\textbf{60.71} &68.88 &76.73 &44.23  &\textbf{55.45} &62.35  &\textbf{56.92} &\textbf{57.80} &12.32 &\textbf{61.10}&  51.35 & 53.65&53.43& 85.75 \\
				& ShortGPT &24.7\%&32.81 	&60.55	&\textbf{71.60}	&\textbf{80.17}	&\textbf{47.13}	 &54.30 &\textbf{62.54} &55.77	&56.41	 &\textbf{15.14}		&60.16 &\textbf{52.11}   &\textbf{58.86} &\textbf{54.43}&\textbf{87.35} \\
				\hline
				\hline
			\end{tabular}
			
		\end{table}
		
		\subsection{Main Results}
		
		To validate the efficacy of our proposed method, we conducted comparative experiments against baseline techniques commonly employed in large language model evaluation. Considering the current structured pruning methods generally reduce parameters by no more than 30\%, we performed experiments with approximately 1/4 of the parameters pruned. The experimental results are presented in Table \ref{tab:llm_comparison_all}. Additional experiments exploring different parameter reduction proportions will be discussed in the subsequent section.
		
		The results demonstrate that the performance of the model pruned by our method significantly surpasses that of the baseline methods, maintaining most of the large language model's capabilities. Furthermore, we note that the approach of reducing the number of layers (ShortGPT/LaCo) outperforms the method of reducing the embedding dimensions (LLMPru./SliceGPT), implying that the model exhibits more redundancy in depth than in width. Further experimental analysis will be presented in the ensuing section.
		
		In Table \ref{tab:llm_comparison_all}, we fully adopted the benchmark, model, and pruning ratio in the LaCo paper. In order to make a more fair comparison with LLMprun. and SliceGPT, we compared them with the same benchmark, model, and pruning ratio in their original paper. The experimental results are shown in Appendix \ref{appendix:fair compare}. Consistent with our findings in Table \ref{tab:llm_comparison_all}, these experiments further demonstrate the significant layer redundancy present in existing large language models, and ShortGPT achieves superior performance compared to other pruning methods.
		
		The results show that coarse-grained pruning methods, such as removing entire layers, often outperform fine-grained approaches like Slice GPT or LLM Pruner.  We speculate that the reason is that the large language model is actually very robust, as shown in Figure \ref{fig:background-redundancy}, removing any deep layer individually actually has very little impact on the final output, which means it is difficult to define the importance of a finer grained module and perform pruning.

		\subsection{Varying  metric and pruning ratio} \label{ana:layerimportance}
		\begin{figure}[h]
			\centering
			\includegraphics[width=0.9\textwidth]{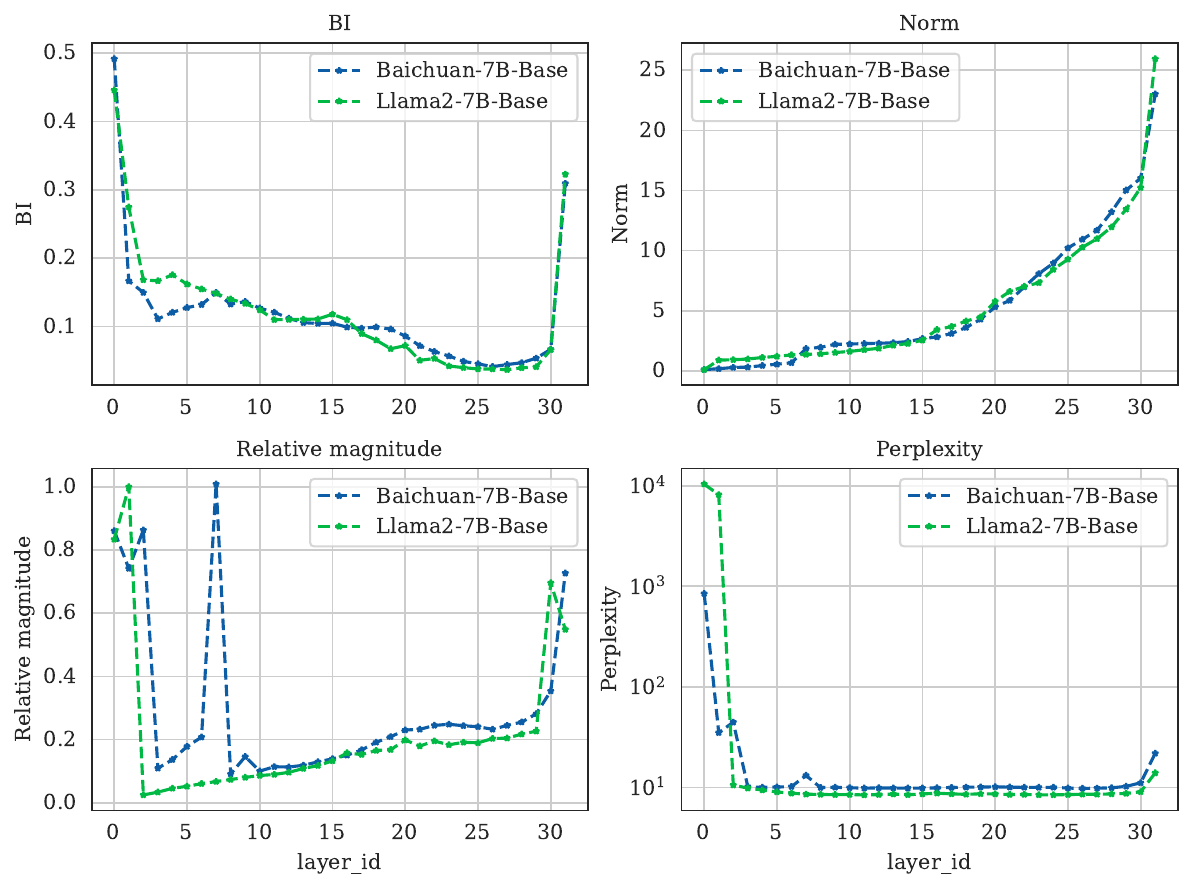}
			\caption{\label{fig:importance-comp}Comparison of different importance metrics. Perplexity is calculated by removing each single layer, other metrics is calculated by hidden states of each layer.}
		\end{figure}
		The core principle of our method is to rank layers by their importance and remove the less significant ones. The choice of importance metric significantly influences the outcome. In this section, we define and compare several different importance metrics:
		\begin{itemize}
			\item \textbf{Sequential}: The importance is directly proportional to the sequence order, with shallower layers being less important. This can be implemented by  assigning the negative value of each layer's index as its importance metric.
			
			\item \textbf{Norm/Reverse-order}: This metric posits that importance is inversely proportional to the sequence order. It assigns higher importance scores to the shallower layers. This method gives the same order as measuring importance by hidden states norm as Figure \ref{fig:importance-comp} shows.
			
			\item \textbf{Relative Magnitude}: Proposed in \cite{samragh2023weight}, this metric assumes layers with larger $ ||\frac{f(x)}{x+f(x)}||$ are of higher importance, where $f$ is the layer transformation function.
			
			\item \textbf{BI}: we calculate the BI score mentioned in Section \ref{method:layerimportacne} as importance metric.
		\end{itemize}

\begin{figure}[t]
			\centering
			\includegraphics[width=0.9\textwidth,height=0.6\textwidth]{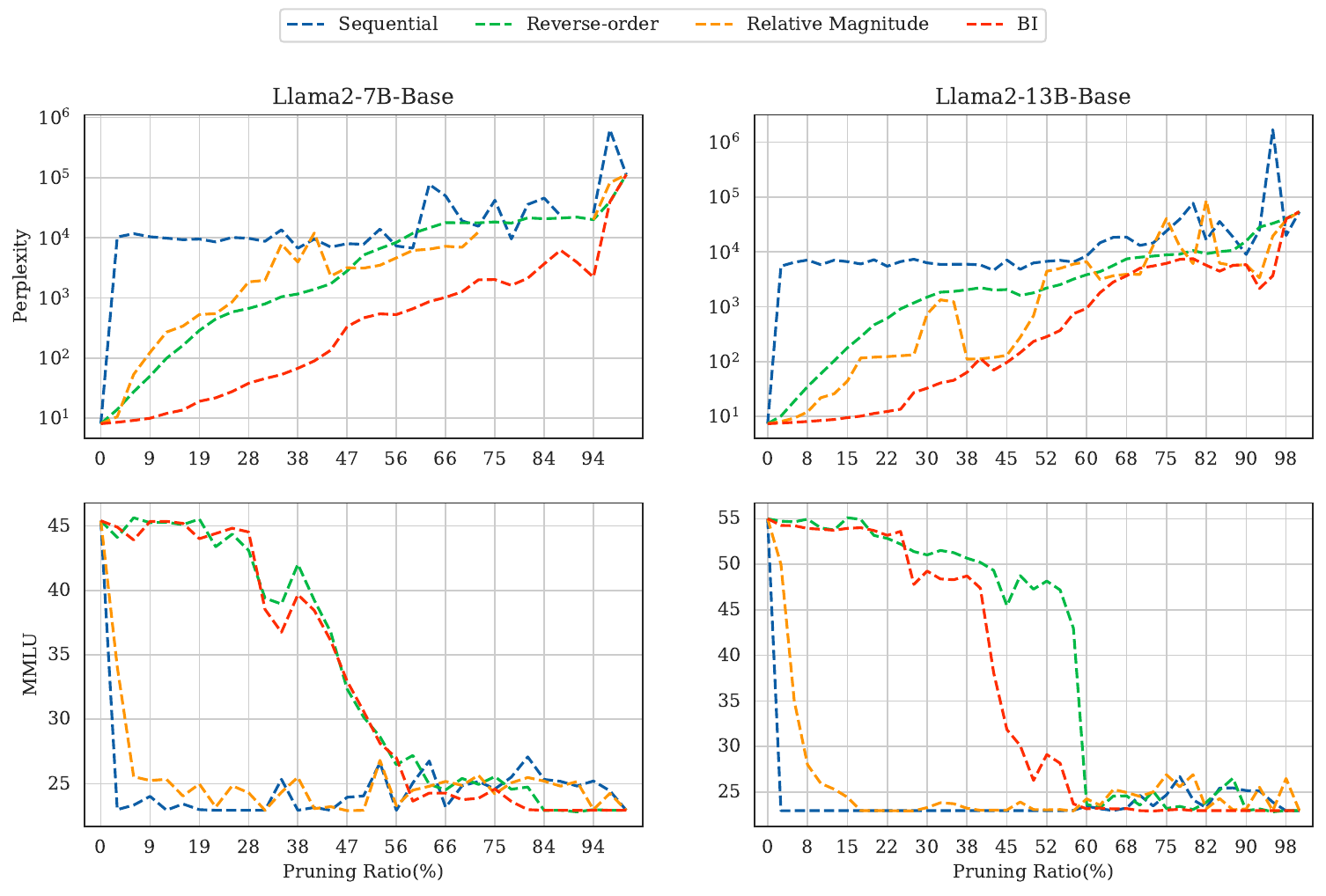}
			\caption{\label{fig:cum-methods-compare} Performance of MMLU and perplexity when we prune by different metrics, with increasing pruning ratio. We can see that as the pruning ratio increases, the performance of the model declines.}
		\end{figure}

		Figure \ref{fig:importance-comp} demonstrates the different metrics. We observe that shallower layers in the LLM network are more crucial than deeper ones. Figure \ref{fig:cum-methods-compare} shows the results of removing layers by different metrics, demonstrating that Our proposed BI outperforms other metrics. The method of Relative Magnitude is highly competitive, indicating that relative values can also reflect the importance to some extent. It is worth noting that the hidden states norm seems to be a good metric when only considering the MMLU benchmark, but the perplexity is relatively poor. 
		
		As a pruning method, we further validated the effects of different pruning ratios on model performance. Experiments were conducted on the Llama2 and Baichuan2 models, observing the Perplexity and MMLU. The results for Llama2, as shown in Figure \ref{fig:cum-methods-compare}, demonstrate that the model's performance generally declines as the pruning ratio increases. However, we observe a notable phenomenon: the MMLU score exhibits a sharp drop at a specific layer. This sudden decrease suggests the presence of certain critical layers within the network that play a particularly important role in maintaining performance. Similar patterns are observed in the Baichuan2 model, as illustrated in Appendix \ref{appendix:details_bc}.

		\subsection{Redundancy on non-transformer LLM}\label{sec:non-transformer}
		
		To determine whether the observed depth redundancy is specific to the Transformer architecture, we extended our investigation to include two popular non-Transformer models,  RWKV-7B\footnote{ We use rwkv-v5-world-7B from https://huggingface.co/RWKV/v5-Eagle-7B-HF} \citep{peng2023rwkv} and Mamba-2.8B  \footnote{We take the model from https://huggingface.co/state-spaces/mamba-2.8b-hf} \citep{gu2023mamba}. Our experiments revealed that these models also exhibit resilience to layer removal, maintaining performance despite the elimination of certain layers. This finding suggests that the redundancy phenomenon may not be unique to Transformer-based models, but rather a common characteristic across current large language models. Table \ref{tab:rwkv_mamba} shows that our method is applicable and effective for both Mamba and RWKV models, suggesting that the redundancy is universal across current LLMs. However, it is worth noting that the RWKV model appears less redundant than Mamba and Transformer models, which warrants further investigation.
		

		\renewcommand\arraystretch{1.3}
		\begin{table}[t]
			\tiny
			\setlength{\tabcolsep}{2.2pt}
			\caption{ShortGPT pruning on RWKV and Mamba.}
			\label{tab:rwkv_mamba}
			\centering
			\begin{tabular}{c|c|ccccccccccccccc}
				\toprule
				Model & Pruning ratio & CMNLI & HeSw & PIQA & CHID & WSC & CoQA & BoolQ & Race-H & Race-M & XSum & C3 & MMLU & CMMLU  &Ave. & Per. \\
				\midrule
				\multirow{5}{*}{Mamba-2.8B}
				& 0\% & 35.97 & 61.84 & 75.52 & 35.56 & 49.69 & 56.35 & 60.67 & 24.9 & 25.3 & 15.03 & 42.08 & 26.29 & 25.32 & 41.12 & 100.00 \\
				& 10.9\% & 32.95 & 59.71 & 73.01 & 32.52 & 49.28 & 52.66 & 51.41 & 24.27 & 25.21 & 14.95 & 41.1 & 26.01 & 25.00 & 39.08 & 95.04  \\
				& 20.3\% & 31.29 & 55.69 & 69.64 & 29.12 & 48.36 & 48.32 & 62.2 & 23.61 & 23.61 & 14.71 & 41.59 & 25.69 & 25.37 & 38.36 & 93.29 \\
				& 25\% & 29.96 & 52.38 & 68.77 & 26.02 & 48.26 & 44.96 & 62.2 & 23.67 & 23.26 & 14.00 & 40.71 & 24.32 & 24.89 & 37.18 & 90.42 \\
				& 31.3\% & 28.25 & 47.02 & 64.91 & 21.38 & 49.69 & 44.96 & 62.17 & 21.87 & 22.77 & 13.77 & 40.44 & 24.48 & 24.77 & 35.59 & 86.55 \\
				\midrule
				\multirow{5}{*}{RWKV-7B}
				& 0\% & 32.07 & 65.98 & 77.09 & 85.36 & 50.00 & 62.65 & 62.72 & 38.56 & 45.47 & 16.5 & 57.97 & 31.85 & 28.54 & 50.37 & 100.00 \\
				& 9.4\% & 32.6 & 56.41 & 73.94 & 78.12 & 50.00 & 49.55 & 62.35 & 25.9 & 25.77 & 9.57 & 54.68 & 27.29 & 25.03 & 43.94 & 87.23 \\
				& 18.8\% & 32.11 & 49.47 & 71.55 & 65.63 & 50.00 & 40.54 & 61.19 & 22.04 & 23.75 & 8.13 & 49.15 & 26.35 & 25 & 40.38 & 80.17\\
				& 25\% & 32.41 & 39.73 & 65.13 & 52.6 & 50.00 & 29.65 & 60.92 & 22.56 & 21.59 & 12.02 & 41.86 & 25.52 & 25.08 & 36.85 & 73.16\\
				& 28.1\% & 33.11 & 32.22 & 60.01 & 32.47 & 50.1 & 28.34 & 60.85 & 22.27 & 21.31 & 10.43 & 37.81 & 25.64 & 25.15 & 33.82 & 67.14\\
				\bottomrule
			\end{tabular}
		\end{table}

		\subsection{Orthogonal to Quantization}
		In this section, we show that our method is orthogonal to quantization methods. We apply our method to Llama2-7B \footnote{We take the model from https://huggingface.co/TheBloke/Llama-2-7B-GPTQ} quantized by GPTQ algorithm. Table \ref{lab:orthogonal} shows that our method is compatible with the quantization-like method. In addition, we compared the performance of applying pruning before quantization \footnote{We use GPTQ algorithm for quantization from https://github.com/AutoGPTQ/AutoGPTQ}. The results shown in the Table \ref{tab:performance_comparison} further indicates that quantization and ShortGPT are orthogonal operations.
		\renewcommand\arraystretch{1.2} 
		\begin{table}[t]
			\small
			\caption{Layer removal results on Llama2-7B-Base-GPTQ.}
			\label{lab:orthogonal}
			\centering
			\begin{tabular}{ccccc}
				\hline
				Model    & Ratio/Layer & Perplexity & MMLU & Throughput (speed up) \\ \hline \hline
				
				Baseline & 0\%/32 &8.03 &43.17  & 4331.23 Token/s (1.00x) \\ \hline
				&  3.1\%/31   & 8.37   &42.88 & 4399.31 Token/s (1.02x)  \\ 
				& 9.4\%/29   &9.44    & 42.31 & 4602.26 Token/s (1.06x)   \\ 
				ShortGPT    & 12.5\%/28   &10.24 &41.62      & 4680.68 Token/s (1.08x)  \\ 
				& 15.6\%/27   &11.42    &43.17  & 4756.94 Token/s (1.10x)   \\ 
				& 25.0\%/24 &22.29  &41.68  & 5045.59 Token/s  (1.16x)  \\ 
				& 27.1\%/23   &40.78&43.35 & 5146.99 Token/s  (1.19x)  \\ 
				\hline
			\end{tabular}
		\end{table}

		\begin{table}[t]
			\small
			\caption{Performance comparison of different methods}
			\label{tab:performance_comparison}
			\centering
			\begin{tabular}{@{}lcc@{}}
				\toprule
				Method & MMLU & CMMLU \\
				\midrule     \midrule
				Llama2-7B-Baseline & 45.4 & 32.9 \\
				\addlinespace
				4-bit quantization & 44.9 & 32.5 \\
				\addlinespace
				Layer removal (27.1\%) & 44.0 & 32.3 \\
				\addlinespace
				4-bit quantization then layer removal & 42.4 & 31.0 \\
				\addlinespace
				Layer removal then 4-bit quantization & 41.2 & 30.5 \\
				\bottomrule
			\end{tabular}
		\end{table}

		\subsection{Post training to restore performance} \label{sec:post training}
		To mitigate the performance loss resulting from layer removal, we explored post-training strategies inspired by \cite{chen2024compressing}. Our approach comprised two key steps: 1)Replacement: We substituted the removed layers with lightweight Multi-Layer Perceptron (MLP) modules. 2)Retraining: We subsequently retrained the modified model. The results in Table \ref{tab:replace} demonstrate the potential of post-train in recover performance loss. Appendix \ref{appendix:post-train} list the training details.
		\begin{table}[htbp]
			\setlength{\tabcolsep}{2.8pt}
			\caption{Post-train Llama2-7B to restore performance.}
			\label{tab:replace}
			\centering
			\tiny
			\begin{tabular}{@{}lccccccccccccccc@{}}
				\toprule
				Method & Avg. & Ratio & CMNLI & HeSw & PIQA & CHID & WSC & CoQA & BoolQ & Race-H & Race-M & XSum & C3 & MMLU & CMMLU \\
				\midrule
				Dense & 47.78 & 0\% & 32.99 & 71.26 &77.91& 41.66 & 50.00 & 64.62 & 71.62 & 35.71 & 34.19 & 19.40 & 43.56 & 45.39&32.92 \\
				ShortGPT & 41.22 & 27.1\% & 32.95 & 53.02 & 66.43 & 24.68 & 52.46 & 47.99 & 	74.41	&32.25	&35.17	&0.67	&39.62	&43.96&	32.25 \\
				ShortGPT+post-train &43.16	&24.0\% & 32.99 &	54.83&	68.12&	31.82&51.37&	58.32	&72.36	&34.18	&34.68&	4.89	&40.37	&44.47	&32.73 \\
				\bottomrule
			\end{tabular}
		\end{table}

		\section{Limitation}\label{limitation}
		Although our method demonstrates strong competitiveness compared to current pruning methods, there are some phenomena that have not been explained. Our experiments reveal that the negative effect of layer removal is more significant on generative tasks compared to multiple-choice tasks. When we remove 25\% layers from Llama2-7B or Baichuan2-7B, the performance in generative tasks such as XSum and C3 deceases to nearly zero, although the performance decline was not as significant on the larger model of the 13B. We speculate that compared to multiple-choice tasks, generative tasks face the problem of accumulated errors and large model is more robust than small one. The reasons behind it still need to be explored. The post-training techniques discussed in Section \ref{sec:post training} have the potential to mitigate this issue and warrant further exploration.

		\section{Related works}
		To reduce the inference cost of large language models and increase their practical applications, there have been many recent works on compressing models, which can be classified into two categories:
		model pruning and quantization. Besides, there are some works aim to study the redundancy of model which is essential for compressing models.
		
		\textbf{Model pruning:} model pruning \citep{lecun1989optimal,han2015learning} is a classic and effective method of reducing model redundancy modules to compress models. The model pruning methods mainly include unstructured pruning and structured pruning. The unstructured pruning simplifies an LLM by removing specific parameters without considering its internal structure, such as SparseGPT \citep{frantar2023massive} and LoRAPrune \citep{zhang2023pruning}. However, this method disregards the overall LLM structure, resulting in an irregular sparse model composition. Another more practical approach is structured pruning, GUM\citep{syed2023prune} makes an analysis of several structured pruning methods for decoder-only LLMs. LLM-Pruner \citep{ma2024llm}  selectively removes non-critical structures according to gradient information. ShearedLLaMA \citep{xia2023sheared} employs targeted structured pruning and dynamic batch loading. LaCo \citep{yang2024laco} used layer merging to compress the model. Compared to the previous method, our method is a simple and efficient structured pruning method.
		
		
		\textbf{Quantization:} quantization \citep{liu2021post,gholami2022survey,dettmers2022llm,dettmers2024qlora} is a widely accepted technique in the field of model compression, which can significantly save the storage and computational costs of deep learning models. Traditional models are generally stored as floating-point numbers, but quantization converts them into integers or other discrete forms. LUT-GEMM \citep{park2022nuqmm} quantifies only weights and optimizes matrix multiplication in LLM using BCQ format. SPQR \citep{dettmers2023spqr}  identifies and isolates abnormal weights, stores them with higher accuracy, and compresses all other weights into 3-4 bits. Our model pruning method and quantization method are orthogonal, which means quantification based on our pruned model can further compress the model.
		
		
		\textbf{Model redundancy:} researchers have long noticed the significant redundancy in nonlinear models \citep{catchpole1997detecting}. In recent years, the transformer model architecture has been widely applied, and researchers have also studied its redundancy. In \citep{bian2021attention}, researchers analyzed redundancy in attention mechanisms, in which clear and similar redundancy patterns (cluster structure) are observed among attention heads. In \citep{dalvi2020analyzing}, researchers dissect two pre-trained models, BERT \citep{devlin2018bert} and XLNet \citep{yang2019xlnet}, studying how much redundancy they exhibit at a representation level and a more fine-grained neuron-level. However, the redundancy in current large language models based on decoder-only structures still needs to be explored. 
		
		\section{Conclusion}
		
		In this work,  we uncovered the significant layer-wise redundancy of LLMs, Our research demonstrates that certain layers contribute minimally to overall network functionality and can be removed without substantially compromising model performance. Based on our observation, We introduce Block influence to quantify the importance of each layer and propose a simple and straightforward pruning method: layer removal. Our experiments demonstrates that it is possible to maintain up to approximately 90\% of a LLM's performance while reducing the model's parameter amount and computational requirements by approximately 25\%. Besides, our method is orthogonal to quantization methods and can be further improved by continual training. We hope that our work can provide some insight for future model compression techniques. 
		Moreover, our work suggests potential avenues for improving the efficiency of model training by reducing inherent redundancy in the future.
		
		\newpage
		\bibliography{baichuan}

\begin{thebibliography}{48}
\providecommand{\natexlab}[1]{#1}
\providecommand{\url}[1]{\texttt{#1}}
\expandafter\ifx\csname urlstyle\endcsname\relax
  \providecommand{\doi}[1]{doi: #1}\else
  \providecommand{\doi}{doi: \begingroup \urlstyle{rm}\Url}\fi

\bibitem[Ashkboos et~al.(2024)Ashkboos, Croci, Nascimento, Hoefler, and
  Hensman]{ashkboos2024slicegpt}
Saleh Ashkboos, Maximilian~L Croci, Marcelo Gennari~do Nascimento, Torsten
  Hoefler, and James Hensman.
\newblock Slicegpt: Compress large language models by deleting rows and
  columns.
\newblock \emph{arXiv preprint arXiv:2401.15024}, 2024.

\bibitem[Bian et~al.(2021)Bian, Huang, Cai, Yuan, and
  Church]{bian2021attention}
Yuchen Bian, Jiaji Huang, Xingyu Cai, Jiahong Yuan, and Kenneth Church.
\newblock On attention redundancy: A comprehensive study.
\newblock In \emph{Proceedings of the 2021 conference of the north american
  chapter of the association for computational linguistics: human language
  technologies}, pp.\  930--945, 2021.

\bibitem[Bisk et~al.(2020)Bisk, Zellers, Gao, Choi, et~al.]{bisk2020piqa}
Yonatan Bisk, Rowan Zellers, Jianfeng Gao, Yejin Choi, et~al.
\newblock Piqa: Reasoning about physical commonsense in natural language.
\newblock In \emph{Proceedings of the AAAI conference on artificial
  intelligence}, pp.\  7432--7439, 2020.

\bibitem[Catchpole \& Morgan(1997)Catchpole and Morgan]{catchpole1997detecting}
Edward~A Catchpole and Byron~JT Morgan.
\newblock Detecting parameter redundancy.
\newblock \emph{Biometrika}, 84\penalty0 (1):\penalty0 187--196, 1997.

\bibitem[Chen et~al.(2024)Chen, Hu, and Zhang]{chen2024compressing}
Xiaodong Chen, Yuxuan Hu, and Jing Zhang.
\newblock Compressing large language models by streamlining the unimportant
  layer.
\newblock \emph{arXiv preprint arXiv:2403.19135}, 2024.

\bibitem[Clark et~al.(2019)Clark, Lee, Chang, Kwiatkowski, Collins, and
  Toutanova]{clark2019boolq}
Christopher Clark, Kenton Lee, Ming-Wei Chang, Tom Kwiatkowski, Michael
  Collins, and Kristina Toutanova.
\newblock Boolq: Exploring the surprising difficulty of natural yes/no
  questions.
\newblock In \emph{Proceedings of the 2019 Conference of the North American
  Chapter of the Association for Computational Linguistics: Human Language
  Technologies, Volume 1 (Long and Short Papers)}, pp.\  2924--2936, 2019.

\bibitem[Dalvi et~al.(2020)Dalvi, Sajjad, Durrani, and
  Belinkov]{dalvi2020analyzing}
Fahim Dalvi, Hassan Sajjad, Nadir Durrani, and Yonatan Belinkov.
\newblock Analyzing redundancy in pretrained transformer models, 2020.

\bibitem[Dettmers et~al.(2022)Dettmers, Lewis, Belkada, and
  Zettlemoyer]{dettmers2022llm}
Tim Dettmers, Mike Lewis, Younes Belkada, and Luke Zettlemoyer.
\newblock Llm. int8 (): 8-bit matrix multiplication for transformers at scale.
\newblock \emph{arXiv preprint arXiv:2208.07339}, 2022.

\bibitem[Dettmers et~al.(2023)Dettmers, Svirschevski, Egiazarian, Kuznedelev,
  Frantar, Ashkboos, Borzunov, Hoefler, and Alistarh]{dettmers2023spqr}
Tim Dettmers, Ruslan Svirschevski, Vage Egiazarian, Denis Kuznedelev, Elias
  Frantar, Saleh Ashkboos, Alexander Borzunov, Torsten Hoefler, and Dan
  Alistarh.
\newblock Spqr: A sparse-quantized representation for near-lossless llm weight
  compression.
\newblock \emph{arXiv preprint arXiv:2306.03078}, 2023.

\bibitem[Dettmers et~al.(2024)Dettmers, Pagnoni, Holtzman, and
  Zettlemoyer]{dettmers2024qlora}
Tim Dettmers, Artidoro Pagnoni, Ari Holtzman, and Luke Zettlemoyer.
\newblock Qlora: Efficient finetuning of quantized llms.
\newblock \emph{Advances in Neural Information Processing Systems}, 36, 2024.

\bibitem[Devlin et~al.(2018)Devlin, Chang, Lee, and Toutanova]{devlin2018bert}
Jacob Devlin, Ming-Wei Chang, Kenton Lee, and Kristina Toutanova.
\newblock Bert: Pre-training of deep bidirectional transformers for language
  understanding.
\newblock \emph{arXiv preprint arXiv:1810.04805}, 2018.

\bibitem[Frantar \& Alistarh(2023)Frantar and Alistarh]{frantar2023massive}
Elias Frantar and Dan Alistarh.
\newblock Massive language models can be accurately pruned in one-shot.
\newblock \emph{arXiv preprint arXiv:2301.00774}, 2023.

\bibitem[Gholami et~al.(2022)Gholami, Kim, Dong, Yao, Mahoney, and
  Keutzer]{gholami2022survey}
Amir Gholami, Sehoon Kim, Zhen Dong, Zhewei Yao, Michael~W Mahoney, and Kurt
  Keutzer.
\newblock A survey of quantization methods for efficient neural network
  inference.
\newblock In \emph{Low-Power Computer Vision}, pp.\  291--326. Chapman and
  Hall/CRC, 2022.

\bibitem[Gu \& Dao(2023)Gu and Dao]{gu2023mamba}
Albert Gu and Tri Dao.
\newblock Mamba: Linear-time sequence modeling with selective state spaces.
\newblock \emph{arXiv preprint arXiv:2312.00752}, 2023.

\bibitem[Han et~al.(2015)Han, Pool, Tran, and Dally]{han2015learning}
Song Han, Jeff Pool, John Tran, and William Dally.
\newblock Learning both weights and connections for efficient neural network.
\newblock \emph{Advances in neural information processing systems}, 28, 2015.

\bibitem[Hasan et~al.(2021)Hasan, Bhattacharjee, Islam, Mubasshir, Li, Kang,
  Rahman, and Shahriyar]{hasan2021xl}
Tahmid Hasan, Abhik Bhattacharjee, Md~Saiful Islam, Kazi Mubasshir, Yuan-Fang
  Li, Yong-Bin Kang, M~Sohel Rahman, and Rifat Shahriyar.
\newblock Xl-sum: Large-scale multilingual abstractive summarization for 44
  languages.
\newblock In \emph{Findings of the Association for Computational Linguistics:
  ACL-IJCNLP 2021}, pp.\  4693--4703, 2021.

\bibitem[Hendrycks et~al.(2020)Hendrycks, Burns, Basart, Zou, Mazeika, Song,
  and Steinhardt]{hendrycks2020measuring}
Dan Hendrycks, Collin Burns, Steven Basart, Andy Zou, Mantas Mazeika, Dawn
  Song, and Jacob Steinhardt.
\newblock Measuring massive multitask language understanding.
\newblock \emph{arXiv preprint arXiv:2009.03300}, 2020.

\bibitem[Hoffmann et~al.(2022)Hoffmann, Borgeaud, Mensch, Buchatskaya, Cai,
  Rutherford, de~Las~Casas, Hendricks, Welbl, Clark, Hennigan, Noland,
  Millican, van~den Driessche, Damoc, Guy, Osindero, Simonyan, Elsen, Rae,
  Vinyals, and Sifre]{hoffmann2022training}
Jordan Hoffmann, Sebastian Borgeaud, Arthur Mensch, Elena Buchatskaya, Trevor
  Cai, Eliza Rutherford, Diego de~Las~Casas, Lisa~Anne Hendricks, Johannes
  Welbl, Aidan Clark, Tom Hennigan, Eric Noland, Katie Millican, George van~den
  Driessche, Bogdan Damoc, Aurelia Guy, Simon Osindero, Karen Simonyan, Erich
  Elsen, Jack~W. Rae, Oriol Vinyals, and Laurent Sifre.
\newblock Training compute-optimal large language models, 2022.

\bibitem[Kaplan et~al.(2020)Kaplan, McCandlish, Henighan, Brown, Chess, Child,
  Gray, Radford, Wu, and Amodei]{kaplan2020scaling}
Jared Kaplan, Sam McCandlish, Tom Henighan, Tom~B. Brown, Benjamin Chess, Rewon
  Child, Scott Gray, Alec Radford, Jeffrey Wu, and Dario Amodei.
\newblock Scaling laws for neural language models, 2020.

\bibitem[Lai et~al.(2017)Lai, Xie, Liu, Yang, and Hovy]{lai2017race}
Guokun Lai, Qizhe Xie, Hanxiao Liu, Yiming Yang, and Eduard Hovy.
\newblock Race: Large-scale reading comprehension dataset from examinations.
\newblock In \emph{Proceedings of the 2017 Conference on Empirical Methods in
  Natural Language Processing}, pp.\  785--794, 2017.

\bibitem[LeCun et~al.(1989)LeCun, Denker, and Solla]{lecun1989optimal}
Yann LeCun, John Denker, and Sara Solla.
\newblock Optimal brain damage.
\newblock \emph{Advances in neural information processing systems}, 2, 1989.

\bibitem[Li et~al.(2024)Li, Zhang, Koto, Yang, Zhao, Gong, Duan, and
  Baldwin]{li2024cmmlu}
Haonan Li, Yixuan Zhang, Fajri Koto, Yifei Yang, Hai Zhao, Yeyun Gong, Nan
  Duan, and Timothy Baldwin.
\newblock Cmmlu: Measuring massive multitask language understanding in chinese,
  2024.

\bibitem[Liu et~al.(2020)Liu, Liu, Gao, Chen, and Han]{liu2020understanding}
Liyuan Liu, Xiaodong Liu, Jianfeng Gao, Weizhu Chen, and Jiawei Han.
\newblock Understanding the difficulty of training transformers.
\newblock In \emph{Proceedings of the 2020 Conference on Empirical Methods in
  Natural Language Processing (EMNLP)}, pp.\  5747--5763, 2020.

\bibitem[Liu et~al.(2021)Liu, Wang, Han, Zhang, Ma, and Gao]{liu2021post}
Zhenhua Liu, Yunhe Wang, Kai Han, Wei Zhang, Siwei Ma, and Wen Gao.
\newblock Post-training quantization for vision transformer.
\newblock \emph{Advances in Neural Information Processing Systems},
  34:\penalty0 28092--28103, 2021.

\bibitem[Ma et~al.(2024)Ma, Fang, and Wang]{ma2024llm}
Xinyin Ma, Gongfan Fang, and Xinchao Wang.
\newblock Llm-pruner: On the structural pruning of large language models.
\newblock \emph{Advances in neural information processing systems}, 36, 2024.

\bibitem[Park et~al.(2022)Park, Park, Kwon, Kim, Lee, and Lee]{park2022nuqmm}
Gunho Park, Baeseong Park, Se~Jung Kwon, Byeongwook Kim, Youngjoo Lee, and
  Dongsoo Lee.
\newblock nuqmm: Quantized matmul for efficient inference of large-scale
  generative language models.
\newblock \emph{arXiv preprint arXiv:2206.09557}, 2022.

\bibitem[Peng et~al.(2023)Peng, Alcaide, Anthony, Albalak, Arcadinho, Cao,
  Cheng, Chung, Grella, GV, et~al.]{peng2023rwkv}
Bo~Peng, Eric Alcaide, Quentin Anthony, Alon Albalak, Samuel Arcadinho, Huanqi
  Cao, Xin Cheng, Michael Chung, Matteo Grella, Kranthi~Kiran GV, et~al.
\newblock Rwkv: Reinventing rnns for the transformer era.
\newblock \emph{arXiv preprint arXiv:2305.13048}, 2023.

\bibitem[Press et~al.(2021)Press, Smith, and Lewis]{press2021train}
Ofir Press, Noah~A Smith, and Mike Lewis.
\newblock Train short, test long: Attention with linear biases enables input
  length extrapolation.
\newblock \emph{arXiv preprint arXiv:2108.12409}, 2021.

\bibitem[Rae et~al.(2019)Rae, Potapenko, Jayakumar, Hillier, and
  Lillicrap]{rae2019compressive}
Jack~W Rae, Anna Potapenko, Siddhant~M Jayakumar, Chloe Hillier, and Timothy~P
  Lillicrap.
\newblock Compressive transformers for long-range sequence modelling.
\newblock In \emph{International Conference on Learning Representations}, 2019.

\bibitem[Reddy et~al.(2019)Reddy, Chen, and Manning]{reddy2019coqa}
Siva Reddy, Danqi Chen, and Christopher~D Manning.
\newblock Coqa: A conversational question answering challenge.
\newblock \emph{Transactions of the Association for Computational Linguistics},
  7:\penalty0 249--266, 2019.

\bibitem[Samragh et~al.(2023)Samragh, Farajtabar, Mehta, Vemulapalli, Faghri,
  Naik, Tuzel, and Rastegari]{samragh2023weight}
Mohammad Samragh, Mehrdad Farajtabar, Sachin Mehta, Raviteja Vemulapalli,
  Fartash Faghri, Devang Naik, Oncel Tuzel, and Mohammad Rastegari.
\newblock Weight subcloning: direct initialization of transformers using larger
  pretrained ones, 2023.

\bibitem[Su et~al.(2024)Su, Ahmed, Lu, Pan, Bo, and Liu]{su2024roformer}
Jianlin Su, Murtadha Ahmed, Yu~Lu, Shengfeng Pan, Wen Bo, and Yunfeng Liu.
\newblock Roformer: Enhanced transformer with rotary position embedding.
\newblock \emph{Neurocomputing}, 568:\penalty0 127063, 2024.

\bibitem[Sun et~al.(2020)Sun, Yu, Yu, and Cardie]{sun2020investigating}
Kai Sun, Dian Yu, Dong Yu, and Claire Cardie.
\newblock Investigating prior knowledge for challenging chinese machine reading
  comprehension.
\newblock \emph{Transactions of the Association for Computational Linguistics},
  8:\penalty0 141--155, 2020.

\bibitem[Syed et~al.(2023)Syed, Guo, and Sundarapandiyan]{syed2023prune}
Aaquib Syed, Phillip~Huang Guo, and Vijaykaarti Sundarapandiyan.
\newblock Prune and tune: Improving efficient pruning techniques for massive
  language models.
\newblock \emph{Arxiv}, 2023.

\bibitem[Touvron et~al.(2023)Touvron, Martin, Stone, Albert, Almahairi, Babaei,
  Bashlykov, Batra, Bhargava, Bhosale, et~al.]{touvron2023llama}
Hugo Touvron, Louis Martin, Kevin Stone, Peter Albert, Amjad Almahairi, Yasmine
  Babaei, Nikolay Bashlykov, Soumya Batra, Prajjwal Bhargava, Shruti Bhosale,
  et~al.
\newblock Llama 2: Open foundation and fine-tuned chat models.
\newblock \emph{arXiv preprint arXiv:2307.09288}, 2023.

\bibitem[Vaswani et~al.(2017)Vaswani, Shazeer, Parmar, Uszkoreit, Jones, Gomez,
  Kaiser, and Polosukhin]{vaswani2017attention}
Ashish Vaswani, Noam Shazeer, Niki Parmar, Jakob Uszkoreit, Llion Jones,
  Aidan~N Gomez, {\L}ukasz Kaiser, and Illia Polosukhin.
\newblock Attention is all you need.
\newblock \emph{Advances in neural information processing systems}, 30, 2017.

\bibitem[Wang et~al.(2024)Wang, Ma, Dong, Huang, Zhang, and
  Wei]{wang2024deepnet}
Hongyu Wang, Shuming Ma, Li~Dong, Shaohan Huang, Dongdong Zhang, and Furu Wei.
\newblock Deepnet: Scaling transformers to 1,000 layers.
\newblock \emph{IEEE Transactions on Pattern Analysis and Machine
  Intelligence}, 2024.

\bibitem[Xia et~al.(2023)Xia, Gao, Zeng, and Chen]{xia2023sheared}
Mengzhou Xia, Tianyu Gao, Zhiyuan Zeng, and Danqi Chen.
\newblock Sheared llama: Accelerating language model pre-training via
  structured pruning.
\newblock \emph{arXiv preprint arXiv:2310.06694}, 2023.

\bibitem[Xiong et~al.(2020)Xiong, Yang, He, Zheng, Zheng, Xing, Zhang, Lan,
  Wang, and Liu]{xiong2020layer}
Ruibin Xiong, Yunchang Yang, Di~He, Kai Zheng, Shuxin Zheng, Chen Xing,
  Huishuai Zhang, Yanyan Lan, Liwei Wang, and Tieyan Liu.
\newblock On layer normalization in the transformer architecture.
\newblock In \emph{International Conference on Machine Learning}, pp.\
  10524--10533. PMLR, 2020.

\bibitem[Xu et~al.(2020)Xu, Hu, Zhang, Li, Cao, Li, Xu, Sun, Yu, Yu,
  et~al.]{xu2020clue}
Liang Xu, Hai Hu, Xuanwei Zhang, Lu~Li, Chenjie Cao, Yudong Li, Yechen Xu, Kai
  Sun, Dian Yu, Cong Yu, et~al.
\newblock Clue: A chinese language understanding evaluation benchmark.
\newblock In \emph{Proceedings of the 28th International Conference on
  Computational Linguistics}, pp.\  4762--4772, 2020.

\bibitem[Yang et~al.(2023)Yang, Xiao, Wang, Zhang, Bian, Yin, Lv, Pan, Wang,
  Yan, et~al.]{yang2023baichuan}
Aiyuan Yang, Bin Xiao, Bingning Wang, Borong Zhang, Ce~Bian, Chao Yin, Chenxu
  Lv, Da~Pan, Dian Wang, Dong Yan, et~al.
\newblock Baichuan 2: Open large-scale language models.
\newblock \emph{arXiv preprint arXiv:2309.10305}, 2023.

\bibitem[Yang et~al.(2024)Yang, Cao, and Zhao]{yang2024laco}
Yifei Yang, Zouying Cao, and Hai Zhao.
\newblock Laco: Large language model pruning via layer collapse, 2024.

\bibitem[Yang et~al.(2019)Yang, Dai, Yang, Carbonell, Salakhutdinov, and
  Le]{yang2019xlnet}
Zhilin Yang, Zihang Dai, Yiming Yang, Jaime Carbonell, Russ~R Salakhutdinov,
  and Quoc~V Le.
\newblock Xlnet: Generalized autoregressive pretraining for language
  understanding.
\newblock \emph{Advances in neural information processing systems}, 32, 2019.

\bibitem[Zellers et~al.(2019)Zellers, Holtzman, Bisk, Farhadi, and
  Choi]{zellers2019hellaswag}
Rowan Zellers, Ari Holtzman, Yonatan Bisk, Ali Farhadi, and Yejin Choi.
\newblock Hellaswag: Can a machine really finish your sentence?
\newblock In \emph{Proceedings of the 57th Annual Meeting of the Association
  for Computational Linguistics}, pp.\  4791--4800, 2019.

\bibitem[Zhang \& Sennrich(2019)Zhang and Sennrich]{zhang2019root}
Biao Zhang and Rico Sennrich.
\newblock Root mean square layer normalization.
\newblock \emph{Advances in Neural Information Processing Systems}, 32, 2019.

\bibitem[Zhang et~al.(2023)Zhang, Shen, Yang, Ou, Yu, Zhuang,
  et~al.]{zhang2023pruning}
Mingyang Zhang, Chunhua Shen, Zhen Yang, Linlin Ou, Xinyi Yu, Bohan Zhuang,
  et~al.
\newblock Pruning meets low-rank parameter-efficient fine-tuning.
\newblock \emph{arXiv preprint arXiv:2305.18403}, 2023.

\bibitem[Zheng et~al.(2019)Zheng, Huang, and Sun]{zheng2019chid}
Chujie Zheng, Minlie Huang, and Aixin Sun.
\newblock Chid: A large-scale chinese idiom dataset for cloze test.
\newblock In \emph{Proceedings of the 57th Annual Meeting of the Association
  for Computational Linguistics}, pp.\  778--787, 2019.

\bibitem[Zhu et~al.(2023)Zhu, Li, Liu, Ma, and Wang]{zhu2023survey}
Xunyu Zhu, Jian Li, Yong Liu, Can Ma, and Weiping Wang.
\newblock A survey on model compression for large language models, 2023.

\end{thebibliography}
		\bibliographystyle{baichuan}
		
		\newpage
		\appendix
		
		\section{Mathematical explanation for why pre-norm brings high similarity}\label{appendix:math}
		We provide a simple explanation here about how pre-norm leads to high deep similarity in this section, here we adopt RMSNorm \citep{zhang2019root} for convenient, which is also the popular pre-norm used in many recent LLMs, such as Llama and Mamba. 
		
		\begin{lemma}\citep{xiong2020layer} \label{lemma:norm}
			At initialization, for the Pre-LN Transformer, $(1 + \frac{L}{2})d \leq \mathbb{E}(||x_{L,i}||^2_2) \leq (1 + \frac{3L}{2})d$ for all $L > 0$ and $i$. Expectations are taken over the input and the randomness of initialization, where the hidden state of $L^{th}$ layer is $x_L$.
		\end{lemma}
		From Lemma \ref{lemma:norm}, the hidden state of the pre-norm model will continuously increase as the number of layers increases. And under the assumption of each component of $x_l$ has a mean of $0$, we can obtain $||x_L||= \Theta (\sqrt{L})$. 
		
		Then we consider $x_{L+1} = x_{L} + f_L(x_L,\theta_{L})$, where $f_L$ is a operation such as Attention or MLP, $\theta_{L}$ is learnable parameters. Then $f_L(x_L,\theta_{L}) = O(1)$ respect to $L$, for Attention as example, $||f_L(x_L,\theta_{L})|| = ||(softmax(Q^TK) X_L/||X_L||\cdot(\sigma_{rms})) W_v W_q|| = O(||\sigma_{rms}||||W_v||||W_o||) = O(1)$ respect to $L$.
		
		Then we can get:
		\begin{align}
			\textit{cos similarity}(X_{L+1},X_{L}) &= \frac{x_{L+1}x_{L}}{||x_{L+1}||||x_{L}||} = \frac{||x_L||^2}{||x_{L+1}||||x_{L}||} +  \frac{f_L(x_L,\theta)x_L}{||x_{L+1}||||x_{L}||}\\
			&\geq \frac{||x_L||^2}{||x_{L+1}||||x_{L}||} - \frac{||f_L(x_L,\theta)||||x_{L}||}{||x_{L+1}||||x_{L}||}\\
			&=\frac{||x_L||}{||x_{L+1}||} - \frac{||f_L(x_L,\theta)||}{x_{L+1}} =\Theta (\sqrt{\frac{L}{L+1})} - O(\sqrt{\frac{1}{L+1}})
		\end{align}
		This means that as the number of layers $L$ increases, the similarity between the input and output of the layer will be high. This means that the role of $f_L$ may be relatively small, and removing it from the network may have a relatively small impact to the model. 
		
		Although the above theoretical analysis is only for randomly initialized models, this phenomenon that deep layer has similar input and output exists in both our own trained models shown in Figure \ref{fig:background-similarity} and existing models in Figure \ref{fig:importance-comp}.
		
		\section{Layer Removal on Baichuan2-series Model}\label{appendix:details_bc}
		
		\begin{figure}[h]
			\centering
			\includegraphics[width=0.6\textwidth]{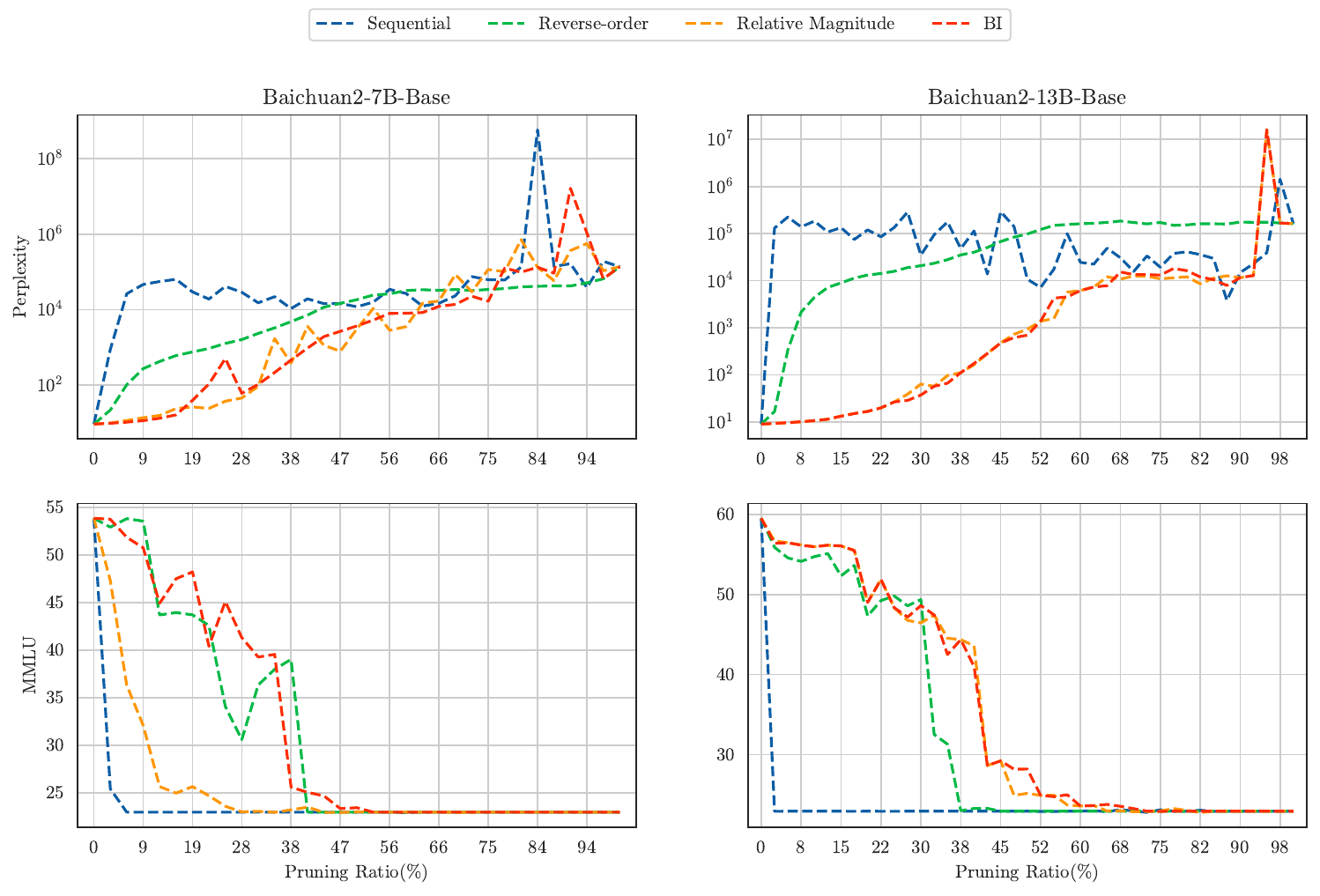}
			\caption{\label{fig:cum-methods-bc}Pruning by different metrics on Baichuan2-series model.}
		\end{figure}

		\newpage
		\section{A Fair comparison with SliceGPT and LLMprun.}\label{appendix:fair compare}
		
		In Table \ref{tab:llm_comparison_all}, we fully adopted the benchmark, model, and pruning ratio in the LaCo's paper. For a fair comparison with LLM pruner and SliceGPT, we do the same experiments in the original paper of LLM pruner and SliceGPT. The results is provided in Table \ref{tab:llmprune_vs_shortgpt}  and Table \ref{tab:slicegpt_vs_shortgpt}. We take \textbf{the same benchmarks, models and pruning ratio} as the corresponding original paper. The results demonstrate that our method is highly competitive.
		
		\begin{table}[h]
			\centering
			\begin{threeparttable}[b]
				
				\tiny
				\caption{ Comparison between ShortGPT and LLM-pruner. The Table is corresponding to the Table 1 of LLM pruner\citep{zhang2023pruning}.}
				\label{tab:llmprune_vs_shortgpt}
				\begin{tabular}{@{}l|ccccccccccc@{}}
					
					\toprule
					\hline
					Model & Pruning ratio & Method & BoolQ & PIQA & Hellaswag & Winogrande & Arc-e & Arc-c & OBQA & Avg. \\  \hline
					\multirow{3}{*}{Llama-7B} & Ratio=0\% & Baseline & 73.18 & 78.35 & 72.99 & 67.01 & 67.45 & 41.38 & 42.4 & 63.25 \\
					& Ratio=20\% & LLM-pruner & 59.39 & 75.57 & 65.34 & 61.33 & 59.18 & 37.12 & 39.80 & 56.82 \\
					& Ratio=21.9 \% & ShortGPT & 68.26 & 72.28 & 61.7 & 63.77 & 60.22 & 39 & 41.6 & 58.12 \\ \hline
					\multirow{3}{*}{Llama-13B} & Ratio=0\% & Baseline & 68.47 & 78.89 & 76.24 & 70.09 & 74.58 & 44.54 & 42.00 & 64.97 \\ 
					& Ratio=20\% & LLM-pruner & 67.68 & 77.15 & 73.41 & 65.11 & 68.35 & 38.4 & 42.4 & 61.79 \\ 
					& Ratio=20\% & ShortGPT & 68.41 & 76.36 & 72.9 & 67.4 & 68.62 & 39.2 & 41 & 61.98 \\ \hline
					\toprule
				\end{tabular}
			\end{threeparttable}
		\end{table}
		\begin{table}[h]
			\centering
			\begin{threeparttable}[b]
				\tiny
				\caption{Comparison between ShortGPT and SliceGPT. The Table is corresponding to the Table 7 of SliceGPT\citep{ashkboos2024slicegpt}.}
				\label{tab:slicegpt_vs_shortgpt}
				\begin{tabular}{@{}l|ccccccccccc@{}}
					\toprule
					\hline
					Model & Pruning ratio & Method  & PIQA & Hellaswag & Winogrande & Arc-e & Arc-c  & Avg. \\  
					\hline
					\multirow{3}{*}{Llama-2-7B} 
					& 0\% & Baseline & 79.11 & 75.99 & 69.06 & 74.58 & 46.25 & 69 \\ 
					\cline{2-9}
					& 20\% & SliceGPT & 71.87 & 58.1 & 63.04 & 69.87 & 43.09 & 63.45 \\
					& 25\% & SliceGPT & 68.55 & 58.1 & 62.04 & 57.46 & 35.07 & 56.15 \\
					& 30\% & SliceGPT & 66.1 & 52.69 & 56.82 & 35.07 & 56.82 & 56.15 \\
					\cline{2-9}
					& 21.9\% & ShortGPT & 72.76 & 66.39 & 66.27 & 59.39 & 39.85 & 60.93 \\
					& 25\% & ShortGPT & 70.53 & 62.68 & 64.7 & 58.39 & 39.51 & 59.16 \\
					& 31.6\% & ShortGPT & 67.87 & 62.19 & 64.38 & 56.57 & 40.86 & 58.37 \\
					\hline
					\multirow{3}{*}{Llama-2-13B} 
					& 0\% & Baseline & 80.47 & 79.39 & 72.22 & 77.48 & 49.23 & 71.76 \\ 
					\cline{2-9}
					& 20\% & SliceGPT & 71.87 & 69.38 & 63.04 & 69.87 & 43.09 & 63.45 \\
					& 25\% & SliceGPT & 68.55 & 67.48 & 58.1 & 62.5 & 37.88 & 58.9 \\
					& 30\% & SliceGPT & 66.1 & 65.11 & 52.69 & 56.82 & 35.07 & 55.16 \\
					\cline{2-9}
					& 20\% & ShortGPT & 76.95 & 74.67 & 71.14 & 69.56 & 45.63 & 67.59 \\
					& 25\% & ShortGPT & 74.39 & 71.65 & 70.98 & 67.09 & 43.93 & 65.61 \\
					& 30\% & ShortGPT & 72.11 & 71.93 & 67.19 & 61.09 & 40.88 & 62.64 \\
					\hline
					\multirow{3}{*}{Llama-2-70B} 
					& 0\% & Baseline & 82.7 & 83.84 & 77.98 & 80.98 & 57.34 & 76.57 \\ 
					\cline{2-9}
					& 20\% & SliceGPT & 76.61 & 72.98 & 74.92 & 80.51 & 55.2 & 72.34 \\
					& 25\% & SliceGPT & 74.92 & 68.74 & 74.92 & 77.9 & 51.71 & 69.75 \\
					& 30\% & SliceGPT & 72.31 & 63.69 & 73.4 & 51.71 & 47.61 & 66.11 \\
					\cline{2-9}
					& 20\% & ShortGPT & 76.02 & 78.87 & 71.69 & 76.02 & 52.95 & 71.68 \\
					& 25\% & ShortGPT & 73.2 & 76.72 & 71.85 & 73.2 & 49.9 & 69.79 \\
					& 30\% & ShortGPT & 74.44 & 75.31 & 72.33 & 74.44 & 49.22 & 69.4 \\
					\hline
					\toprule
				\end{tabular}
			\end{threeparttable}
		\end{table}

		\section{Detailed Strategies for Layer Removal}\label{appendix:remove_strategy}
		We list the details of different layer removal strategies in Table \ref{tab:strategy-detail}. The concrete removed layers by ShortGPT in Table \ref{tab:llm_comparison_all} are listed in Table \ref{tab:benchmark-removed-layers.}
		
		\begin{table}[h]
			\centering
			\small
			\caption{Setup of Removed Layers for Benchmark Models.}
			\label{tab:benchmark-removed-layers.}
			\begin{tabular}{@{}lc@{}} 
				\toprule
				Model & Removed Layers \\
				\midrule
				Llama-2-7B & 27, 26, 25, 28, 24, 29, 23, 21, 22 \\
				Llama-2-13B & 33, 31, 32, 30, 29, 34, 28, 35, 27, 26 \\
				Baichuan-2-7B & 26, 27, 25, 28, 24, 29, 23, 22, 30 \\
				Baichuan-2-13B &32, 31, 33, 30, 34, 29, 28, 35, 27, 26 \\
				\bottomrule
			\end{tabular}
		\end{table}
		
		\begin{table}[h]
			
			\centering
			\caption{Strategies for Layer Removal in Models.}
			\label{tab:strategy-detail}
			\small
			\begin{tabular}{@{}lp{8cm}@{}} 
				
				\toprule
				Strategy & Description \\
				\midrule
				Sequential & Layers are removed sequentially from the beginning of the model. The process starts with layer 0 and progressively includes more layers for removal (e.g., \{0\}, \{0, 1\}, \ldots). \\
				\hline
				Reverse-order & This strategy involves starting from the model's final layer and progressively removing layers in reverse order (e.g., \{-1\}, \{-1, -2\}, \ldots). \\
				\hline
				Relative Magnitude & Layers are removed in ascending order based on their Relative Magnitude values. The removal process accumulates layers from those with the smallest to the largest values, mirroring the sequential strategy's accumulation method. \\
				\hline
				BI (Block Influence) & Follows a similar accumulation approach as the Sequential strategy, but layers are ordered and removed according to their BI values, starting from the lowest and moving to the highest. \\
				\bottomrule
			\end{tabular}
		\end{table}
		
		\section{Setup for training post-norm model and 
			pre-norm model}\label{appendix:post and pre}
		
		We have listed the specific training settings for pre norm and post norm in Table \ref{tab:params}.
		
		\begin{table}[h]
			\centering
			\tiny
			\caption{Training Parameters.}
			\label{tab:params}
			\begin{tabular}{ll} 
				\toprule
				\textbf{Parameter} & \textbf{Value} \\ \midrule
				Global Batch Size & 2048 \\
				Sequence length & 4096 \\
				Precision & bf16 \\
				Learning Rate Scheduler & cosine \\
				Max Learning Rate  & 4e-4 \\
				Min Learning Rate  & 5e-5 \\
				Warm-up steps & 3000 \\
				Training Tokens & 200B \\
				Weight Decay  & 0.1 \\
				Adam Beta1 & 0.9 \\
				Adam Beta2 & 0.98 \\
				Gradient Clip & 1.0 \\ 
				Tokenizer & Llama2 \\
				Layers & 32 \\
				Hidden state & 2048 \\
				Attention heads & 32\\
				Head dim & 64 \\
				FFN size & 5504 \\
				Activation function & Silu 
				\\\bottomrule
			\end{tabular}
		\end{table}
		
		\section{post-training settings}\label{appendix:post-train}
		We replace the removed layer with a lightweight gated MLP layer with hidden size = 2048. Table \ref{appendix:tab:post-train} show the post training settings.
		\begin{table}[h]
			\centering
			\tiny
			\caption{Post-training Parameters.}
			\label{appendix:tab:post-train}
			\begin{tabular}{ll} 
				\toprule
				\textbf{Parameter} & \textbf{Value} \\ \midrule
				Global Batch Size & 2048 \\
				Sequence length & 4096 \\
				Precision & bf16 \\
				Learning Rate Scheduler & cosine \\
				Max Learning Rate  & 2e-5 \\
				Min Learning Rate  & 1e-5 \\
				Warm-up steps & 3000 \\
				Training Tokens & 50B \\
				Weight Decay  & 0.1 \\
				Adam Beta1 & 0.9 \\
				Adam Beta2 & 0.98 \\
				Gradient Clip & 1.0
				\\\bottomrule
			\end{tabular}
		\end{table}

		\newpage

		
			
		
		\newpage
		
		
		
		\section{Evaluation Benchmarks} \label{appendix:benchmark}
		In order to comprehensively evaluate the changes in the ability of large language models before and after pruning, we conducted evaluations on the most commonly used Benchmark MMLU \cite{hendrycks2020measuring}, CMMLU \cite{li2024cmmlu}  for evaluating large models. In addition, we also followed LaCo \cite{yang2024laco} to evaluate a wider dataset. 
		
		\textbf{MMLU} \cite{hendrycks2020measuring} is a benchmark aimed at measuring the knowledge acquired during pre-training by specifically evaluating models in zero-shot and few-shot settings. This makes benchmarks more challenging and similar to the way we evaluate humans. This benchmark covers 57 subjects including STEM, humanities, social sciences, etc. Its difficulty ranges from beginner to advanced professional level, and it tests world knowledge and problem-solving ability.
		
		\textbf{CMMLU} \cite{li2024cmmlu} is a comprehensive Chinese language assessment dataset designed specifically to evaluate LLM's advanced knowledge and reasoning abilities in the context of Chinese language and culture. CMMLU covers 67 topics, from elementary school to university or professional level. Including natural sciences, as well as humanities and social sciences, it also includes many contents with Chinese characteristics.
		
		
		\textbf{CMNLI} \cite{xu2020clue}  is part of the Chinese language understanding assessment benchmark. It consists of two parts: XNLI and MNLI. \textbf{HellaSwag (HeSw)} \cite{zellers2019hellaswag}  is a challenging dataset for evaluating commonsense NLI that is especially hard for state-of-the-art models, though its questions are trivial for humans. \textbf{PIQA} \cite{bisk2020piqa} is a multi-choice question and answer dataset that focuses on daily scenarios. This dataset explores the model's grasp of the laws of the real physical world through daily scenarios. \textbf{CHID} \cite{zheng2019chid} is an idiom cloze test dataset that mainly focuses on the selection of candidate words and the representation of idioms. \textbf{CoQA} \cite{reddy2019coqa} is a large-scale dataset used for conversational question-answering tasks, containing over 127000 questions and their corresponding answers. \textbf{BoolQ} \cite{clark2019boolq} is a question-answer dataset containing 15942 examples of yes/no questions. These problems occur naturally - they are generated in an environment that is silent and unconstrained. \textbf{Race} \cite{lai2017race} is a large-scale reading comprehension dataset collected from English examinations in China, which are designed for middle school and high school students. \textbf{XSum}\cite{hasan2021xl} is used to evaluate abstract single document summarization systems. The goal is to create a short, one-sentence new summary of what the article is about. \textbf{C3} \cite{sun2020investigating} is a machine reading comprehension dataset with multiple choices, consisting of multiple-choice questions, reading materials from Chinese proficiency exams, and ethnic Chinese exams. \textbf{PG19} \cite{rae2019compressive}  is a long document dataset from books used to test the effectiveness of language modeling. 
		
		\section{Hardware Environment}\label{hard.env}
		The platform we use to experiment is GPU heterogeneous platform.
		The hardware of our platform is shown in Table \ref{tab:hardware} 
		\begin{table}[h]
			\centering
			\caption{Setup of Removed Layers for Benchmark Models.}
			\label{tab:hardware}
			\begin{tabular}{@{}lc@{}} 
				\toprule
				Name & Details  \\
				\midrule
				CPU& 2x Intel(R) Xeon(R) Gold 6430 CPU @ 2.1GHz\\
				GPU& 8x NVIDIA A100-80GB Tensor Core GPU\\
				\bottomrule
			\end{tabular}
		\end{table}

		\end{document}